\documentclass[10pt,twocolumn,letterpaper]{article}

\usepackage[pagenumbers]{cvpr} % To force page numbers, e.g. for an arXiv version

\usepackage{graphicx}
\usepackage{amsmath}
\usepackage{amssymb}
\usepackage{booktabs}
\usepackage{bm}
\usepackage{wrapfig}
\usepackage{lipsum}
\usepackage{wrapfig}
\usepackage{array,multirow}

\usepackage[pagebackref,breaklinks,colorlinks]{hyperref}

\usepackage[capitalize]{cleveref}
\crefname{section}{Sec.}{Secs.}
\Crefname{section}{Section}{Sections}
\Crefname{table}{Table}{Tables}
\crefname{table}{Tab.}{Tabs.}

\def\cvprPaperID{2968} % *** Enter the CVPR Paper ID here
\def\confName{CVPR}
\def\confYear{2022}

\begin{document}

%%%%%%%%% TITLE - PLEASE UPDATE
\title{Self-supervised 3D Human Mesh Recovery from Noisy Point Clouds}

\author{ Xinxin Zuo{$^{\ast \S}$}  \qquad Sen Wang{$^{\ast \S}$}  \qquad  Qiang Sun{$^\dagger$} \qquad Minglun Gong{$^\S$} \qquad  Li Cheng{$^\ast$}\\
{$^\ast$}University of Alberta \quad {$^\dagger$}University of Toronto  \quad {$^\S$}University of Guleph\\
%Institution1 address\\
{\tt\small xzuo, sen9,lcheng5@ualberta.org}
% For a paper whose authors are all at the same institution,
% omit the following lines up until the closing ``}''.
% Additional authors and addresses can be added with ``\and'',
% just like the second author.
% To save space, use either the email address or home page, not both
%\and
%Second Author\\
%Institution2\\
%IFirst line of institution2 address\\
%{\tt\small secondauthor@i2.org}
}
\maketitle

%%%%%%%%% ABSTRACT
\begin{abstract}
This paper presents a novel self-supervised approach to reconstruct human shape and pose from noisy point cloud data. Relying on large amount of dataset with ground-truth annotations, recent learning-based approaches predict correspondences for every vertice on the point cloud; Chamfer distance is usually used to minimize the distance between a deformed template model and the input point cloud. 
However, Chamfer distance is quite sensitive to noise and outliers, thus could be unreliable to assign correspondences. To address these issues, we model the probability distribution of the input point cloud as generated from a parametric human model under a Gaussian Mixture Model. 
Instead of explicitly aligning correspondences, we treat the process of correspondence search as an implicit probabilistic association by updating the posterior probability of the template model given the input. A novel self-supervised loss is further derived which penalizes the discrepancy between the deformed template and the input point cloud conditioned on the posterior probability. Our approach is very flexible, which works with both complete point cloud and incomplete ones including even a single depth image as input. 
Compared to previous self-supervised methods, our method shows the capability to deal with substantial noise and outliers. Extensive experiments conducted on various public synthetic datasets as well as a very noisy real dataset (i.e. CMU Panoptic) demonstrate the superior performance of our approach over the state-of-the-art methods. 
\end{abstract}

%Our network is trained from scratch with no need to warm-up the network with supervised data.

%%%%%%%%% BODY TEXT
\section{Introduction}
\begin{figure*}[ht]
    \centering
    \includegraphics[width=0.85\textwidth]{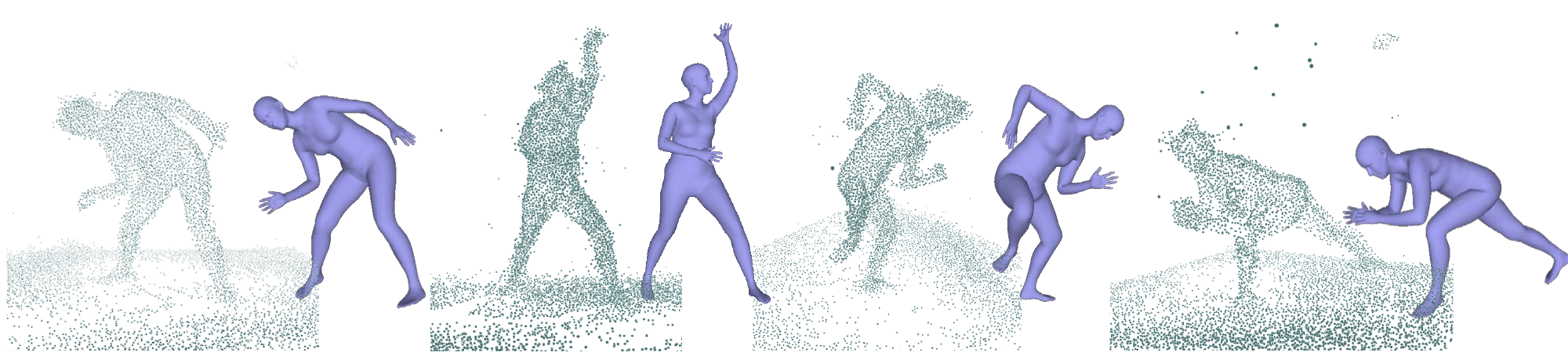}
    \caption{Sampled results of our reconstructed human mesh models from noisy point clouds.}
    \vspace{-7pt}
    \label{fig:titleImage}
\end{figure*}

With the rapid development of sensing technology, it becomes increasingly popular to digitize human in 3D scans~\cite{dou2016fusion4d,yu2018doublefusion,li2021posefusion}. This gives rise to a surging demand for techniques to edit, control and animate the acquired 3D content~\cite{saito2021scanimate,ma2021power,ma2021scale} which often involves turning a 3D human scan into a parametric rigged representation. 
Meanwhile, it is usually difficult to annotate point clouds captured in real life. 
This motivates us to investigate in this paper a self-supervised approach to fit a parametric rigged representation into an input human scan. Without loss of generality, we focus on fitting a SMPL model~\cite{loper2015smpl} to a 3D human point cloud.

Traditional approaches address this parametric model based fitting problem using iterative closest point (ICP) strategies~\cite{besl1992icp,pons2017clothcap}. The main idea is to iteratively search for the correspondences between the parametric model and the input scan, then re-estimate the parameters involving the pose, shape and non-rigid surface displacement of the template model. They however heavily rely on a good initialization to avoid being stuck into a local minimum solution. % due to large pose and shape variations. 
Existing learning based approaches~\cite{wei2016dense,litany2017deepfunctional,ginzburg2020cyclic}, on the other hand, typically rely on well-annotated training set, being synthetic or real-world datasets. More importantly, the input point clouds are assumed to have relatively clean surfaces. They are nonetheless difficult to deal with practical scenarios, where it is often difficult to obtain 3D shape annotations, and outliers are prevalent in the input scans. 
Last but not least, existing works usually requires as input the presence of a complete point cloud from full-view 3D scans. This severely limits their use when only incomplete point cloud from partial-view (e.g. a depth image) is available, a more common scenario in practice.

These observations motivate us to propose an self-supervised probabilistic-based 3D fitting approach that can cope with noisy and partial input scans.
The key idea is instead of directly regressing the one-to-one correspondence~\cite{bhatnagar2020loopreg,bhatnagar2020IPnet} from the input point cloud to the template model, we model the probability of the vertices on the input point cloud with a Gaussian Mixture Model (GMM) where the centroids of GMM are the vertices on the human template. In addition, we introduce a probability for the input vertices of being outliers. Specifically, our network takes the point cloud as input and predict the human shape and pose parameters of the SMPL model. A probabilistic correspondence association module is proposed to update the soft correspondences between the current predict SMPL model and the input scan in each training iteration. Then we define our self-supervised loss function which minimizes the distance between the input scan and the parametric human model with the established soft correspondences. The network gets updated through back-propagation with the updated training loss. 

The contributions of this paper are summarized: 1) We propose a novel self-supervised method to reconstruct human shape and pose from noisy point cloud; 2) We propose a probabilistic loss function derived from GMM and encode the correspondences in the probability distribution which is naturally differential. It works on both complete and incomplete point cloud and is robust to outliers; 3) We evaluate the proposed method on several public datasets and outperform the current state-of-the-art methods. And we also demonstrate the effectiveness of our method in real captured noisy point cloud. We show some sampled results in Fig.~\ref{fig:titleImage}.

% The contributions of this paper are summarized:
% \begin{itemize}
%     \item We propose a novel unsupervised method to reconstruct human shape and pose from point cloud. The network can be trained in a fully unsupervised manner with no need to warm-up the network with supervised data.
%   \item We propose a probabilistic loss function derived from GMM and encode the correspondences in the probability distribution which is naturally differential. It works on both complete and incomplete point cloud and is robust to outliers.
%   \item We evaluate the proposed method on several public datasets and outperform the current state-of-the-art methods. And We also demonstrate the effectiveness of our method in real captured noisy point cloud.
% \end{itemize}
%Parametric model

%SMPL~\cite{loper2015smpl}

%SMPLIfy-x~\cite{pavlakos2019smplifyx}
%STAR~\cite{Osman2020star}
\section{Related Work}
\subsection{Model-based Surface Fitting}
Traditional approaches deal with the model fitting of 3D scans of articulated, highly non-planar objects like human bodies~\cite{Bogo2014faust,bogo2017dynamicfaust} or human hands~\cite{romero2017embodiedMANO,wan2019self} using nonrigid ICP based techniques~\cite{besl1992icp,pons2017clothcap} where the correspondences matching and parametric model fitting are conducted iteratively. To account for the large pose displacement, researchers proposed to use sliding correspondences~\cite{li2008global} or probabilistic correspondence association~\cite{ye2014real,horaud2010rigid} for more effective correspondence matching. However, they generally require a good model initialization, otherwise it will easily fall into local minimum. There are also approaches that focus on surface tracking over the 3D video sequence~\cite{yu2018doublefusion,pons2017clothcap}, in which case they usually have a predefined starting pose.

%With full supervision of the groundtruth human models, there are also efforts that directly predict the latent parameters of the human model~\cite{jiang2019skeleton} or human model after deformation~\cite{groueix20183dcoded,wang2021locallyPTF,wang2020sequential}.

Nowadays, to automate the model fitting process and eliminate the dependency on initialization, researchers take advantage of the deep learning techniques and train a neural network to predict the parametric model from input point cloud with the supervision of the groundtruth human models~\cite{li2019lbsself,jiang2019skeleton,groueix20183dcoded}. More recently, implicit functions~\cite{bhatnagar2020IPnet,wang2021locallyPTF,chibane2020implicit,mihajlovic2021leap} were proposed and were combined with the parametric model to restore the human surface details. For example, in paper~\cite{wang2021locallyPTF}, they propose to learn a set of functions to map any query point in posed space to its correspond position in rest-pose space and conduct occupancy classification in the unposed-space to finally restore the human shape. Another branch of model fitting is to deal with this problem in two steps: they predict dense correspondences between the template model and the input point cloud~\cite{wang2020sequential,bhatnagar2020loopreg,marin2020farm,wang2021learning}, from which the parameters of the template model can be optimized. We will review the related works on correspondence matching in the following subsection.
In addition to model based surface fitting, there are also efforts on general non-rigid surface registration by predicting the surface deformation~\cite{feng2021recurrent} where the problem of topology changes would be difficult to resolve. 
%LBS autoencoder~\cite{li2019lbsself} 
%3d coded~\cite{groueix20183dcoded}
%Functional Automatic Registration Method for 3D Human Bodies~\cite{marin2020farm}
%Loopreg~\cite{bhatnagar2020loopreg}
%Locally PTF~\cite{wang2021locallyPTF}
%IPNet~\cite{bhatnagar2020IPnet}
%Skeleton-Aware~\cite{jiang2019skeleton}
%Sequential~\cite{wang2020sequential}

%As another typical articulated model, human hand is modeled where similar model fitting techniques are proposed~\cite{wan2019self} with extra effort on dealing with intensive interaction and intrusion. Detailed review can be found in paper.

%HMR~\cite{kanazawa2018hmr}
%EM~\cite{myronenko2010cpd}
%learning clothed people~\cite{alldieck2019learningcloth, corona2021smplicit}
\subsection{Correspondence}
%Dense non-rigid shape correspondence such as~\cite{chen2015robust} focuses on establish correspondences between the template model and the input point cloud, a related problem that is critical for many downstream tasks including pose or texture transfer across surfaces. 
Traditional methods on finding shape correspondences deal with the matching problem using handcrafted local shape descriptors~\cite{aubry2011wave,bronstein2010scale,tombari2010unique} which was supposed to maintain invariance under a wide class of transformations the shape can undergo. There are also efforts on finding globally consistent sets of maps~\cite{nguyen2011optimization,kim2011blended} between shapes. Operating in a low-dimension space composed of the Laplace-Beltrami basis, the Functional Maps~\cite{ovsjanikov2012functional} reduced dimensionality of the problem drastically by converting the point-level correspondence to the function-level correspondence.
%Taking the global consistency into consideration, some methods concentrated on measuring and optimizing consistency of sets of maps, among which the functional maps have the most impact.

To secure reliable correspondences, learning based methods were developed by training a neural network to predict the dense correspondences~\cite{wei2016dense,litany2017deepfunctional,ginzburg2020cyclic,wang2020sequential} between template model and input point cloud.
%between template model and input point cloud, using synthetic or real-world datasets.
%Learning based methods train a neural network to predict the dense correspondences between the template model and the input point cloud~\cite{bhatnagar2020loopreg}. 
%Some works tackle this problems in two-steps: correspondences matching and model fitting. 
For instance, targeting at human bodies, Wei~\cite{wei2016dense} trained a feature descriptor on depth map pixels and treated the correspondence matching as a body region classification problem. Recently Deep Functional Maps~\cite{marin2020farm,litany2017deepfunctional,roufosse2019unsupervised,ginzburg2020cyclic} were developed to compute correspondences across 3D shapes while optimizing for global structural properties of the surface. Another widely-used relaxation for matching problems~\cite{eisenberger2020deep,solomon2016entropic,vestner2017product}, Optimal Transport, rely on large-scale dense matrices (e.g. geodesic distances or heat kernels) for surface matching. 
More recently, a self-supervised method called Loopreg~\cite{bhatnagar2020loopreg} was proposed with a differential registration loop to predict correspondences and register template models to the input point cloud.
%In addition to supervised approaches, paper \cite{bhatnagar2020loopreg} form a registration loop for the correspondences searching and model fitting and handle the model fitting problem in a self-supervised way.

Although the learning based methods have been widely explored on model fitting as well as correspondence matching, they usually assume clean input scan and compute one-to-one correspondence between surfaces. But the assumption can not hold for real captured data when the input point cloud is quite noisy with outliers. On the other hand, a large dataset with groundtruth human model annotations is also required to train the network. On the contrary, our proposed approach can work in a fully self-supervised manner and is also robust to outliers with our probabilistic correspondence association.

\section{Our Approach}
\label{sec:method}
To reconstruct human surface from point cloud, we follow the template based surface fitting strategy. Therefore, our goal is to optimize the parameters in the human model so that the deformed human model can match the input point cloud.

%Basically, there are two key components for the model based surface fitting problem: correspondence searching and parametric model optimization with the correspondences. 
The traditional approaches solve the problem in an iterative way through correspondence search and model fitting.
%Starting from an initial model estimation, they first compute the correspondences between the template model and the input point cloud and the template gets updated with the correspondences afterwards.
Mathematically, the human model is reconstructed by minimizing the following objective function
\begin{equation}
L(\Theta,\bm{c}) = \sum_{n=1}^{N} dist \big( \bm{v}_n, M(\bm{c}_m, \Theta)) \big),
\vspace{-3pt}
\end{equation}
where $\bm{c}_m$ denotes the correspondences in the human model for $\bm{v}_n$ from input point cloud which are usually computed via nearest search, and the human model($M(\Theta)$) gets updated by minimizing the distance function $dist(\cdot)$ which can be point-to-point or point-to-surface distance between the correspondences. However, the optimization will fall into local minimal especially when the initial model is far away from the input. 
%From the perspective of neural networks, the network is trained to predict the correspondences and the parameters of the template model. It is rather difficult to train the network to predict correspondences in an unsupervised manner and the only constraint is to penalize the discrepancy between the deformed template model and the input point cloud. 

Generally, the above distance metrics or Chamfer loss are also used as an unsupervised loss to train the network for model registration~\cite{bhatnagar2020loopreg,li2019lbsself}.
%For example, recently in paper~\cite{bhatnagar2020loopreg}, they create a differentiable registration loop and regress the correspondences by enforcing a point-to-surface distance loss between the deformed template model and the input point cloud.  
However, similar to the local minimal in traditional model fitting approaches, it is hard for the network to converge if trained with the above distance loss without using supervised data to warm-up the network. More importantly, this distance function is sensitive to outliers. When we consider the correspondence association, the previous methods usually predict correspondences for every vertex of the input point cloud, but it is not appropriate to assign correspondences for vertices belonging to outliers.

To resolve the above issues, instead of explicitly assigning correspondences, we propose a probabilistic correspondence association module and define a novel distance function as the self-supervised fitting loss which can be differential and robust to outliers. The proposed loss function allows us to train the network from scratch. In the following section, we will describe the derivation of our loss function as well as the network structure in detail.

\subsection{Probabilistic Human Model}
We model the probability of the input point cloud with a Gaussian Mixture Model. Different from previous works~\cite{browne2011model,biggs2020left} where a single GMM model is used to model the distribution of the overall input, in this paper we propose to model the distribution of every vertex on the input point cloud with a GMM. Specifically, for each vertice of the input point cloud, it is assumed to be generated by a GMM whose centroids are the vertices of the deformed template model. To further compensate for the outliers in the input point cloud, we use a uniform distribution to define the probability of the vertices being outliers. Mathematically, the probability of each vertex of the input scan $\bm{v}_n$ can be expressed as
\begin{equation}
\begin{split}
p(\bm{v}_n)  & = (1-\mu) \sum_{m=1}^{|M|} p(m) p(\bm{v}_n|m) + \mu\frac{1}{N}  \\ & = (1-\mu) \sum_{m=1}^{|M|} \pi_{mn} \mathcal{N}(\bm{v}_n|m) +  \mu\frac{1}{N},
\end{split}
\end{equation}  
where $N$ and $|M|$ are the number of vertices of the input point cloud and the template human model respectively. $\mathcal{N}(\bm{v}_n|m)$ is a normal distribution and $\mu$ is the approximation for the percentage of outliers, which is considered to be evenly distributed. $\pi_{mn}$ represents the coefficients of the mixture model which can be viewed as the probability of assigning vertice $\bm{v}_n$ in the input point cloud and the vertice $M_m$ on the template human model as correspondence.
% \begin{equation}
% \begin{aligned}
% p(\bm{v_n}) & = (1-\mu) \sum_{m=1}^{M} p(m) p(\bm{v_n}|m) +  \mu\frac{1}{N} \\
% & = (1-\mu) \sum_{m=1}^{M} \pi_{mn} \mathcal{N}(\bm{v_n}|m) +  \mu\frac{1}{N}
% \end{aligned}
% \end{equation}  

\begin{figure*}[!ht]
    \centering
    \includegraphics[width=0.9\textwidth]{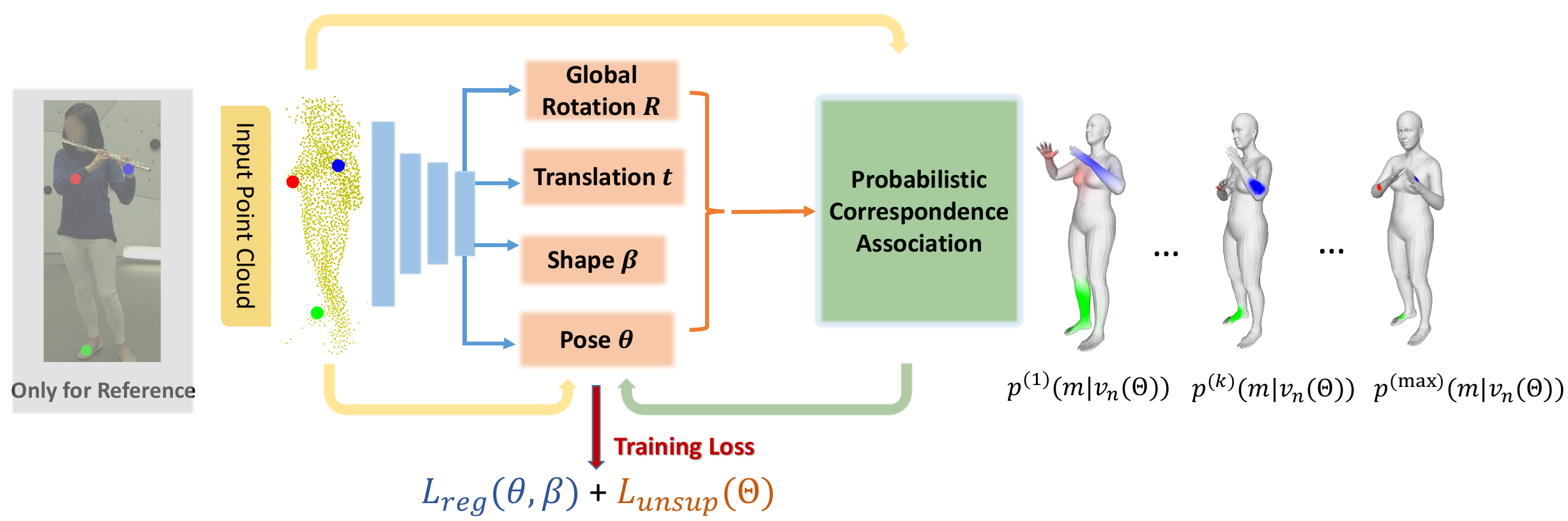}
    \caption{Overall structure. The detailed explanation is in Sec.~\ref{sec:net}. \textit{The color image is not taken as input but only for better visualization.}}
    \label{fig:pipeline}
\end{figure*}

We use equal isotropic covariances $\sigma^2$ for the Gaussian model and therefore the conditional distribution $p(\bm{v}_n|m)$ is formulated as
\begin{equation}
\label{eq:GMM}
p(\bm{v}_n|m) = \frac{1}{(2\pi\sigma^2)^{\frac{\bm{3}}{2}}}  \exp \left(-\frac{||M_m(\Theta) -\bm{v}_n||^2}{2\sigma^2} \right).
\end{equation}
In the above conditioned probability function, $M(\Theta)$ is the parametric human model with the latent parameters denoted as $\Theta$. The proposed techniques can apply generally to any 3D statistical model and in this paper, we have used the SMPL human model~\cite{loper2015smpl} which is parameterized by the shape $\beta$ and pose $\theta$ coefficients. Therefore, the parameters of the human model are $\Theta=\{ \theta, \beta, R,  t\}$ where $R$ and $t$ are the global rotation matrix and translation vector to transform the template human model to the input point cloud. Instead of using Euler angles to represent the global rotation~\cite{bhatnagar2020IPnet,bhatnagar2020loopreg}, in this paper we employ the 6D rotation representation which has been proved to be continuous in real Euclidean spaces and more suitable for learning~\cite{zhou2019continuity, xu2019disn}.
Given those parameters, the human model is generated as:
\begin{equation}
M(\Theta) = R* W(T_{P}(\beta, \theta), J(\beta), \theta, \Omega, T_0) + t
\end{equation}
\begin{equation}
T_{P}(\bm{\beta}, \bm{\theta}) = T_0 + B_{S}(\bm{\beta}) + B_{P}(\bm{\theta})
\end{equation}
where $T_0$ is the base template mesh, $B_{S}(\bm{\beta})$ and $B_{P}(\bm{\theta})$ are vectors of vertices representing offsets from the base template as controlled by the shape and pose parameters respectively. Therefore, $T_p$ is the mesh of base template with the addition of both shape $B_{S}(\bm{\beta})$ and pose blend shapes $B_{P}(\bm{\theta})$. $J(\bm{\beta})$ is the joints position under the rest pose as controlled by the shape parameters. $W()$ is a blend skinning function which transforms the mesh from $T$ pose to the current pose $\bm{\theta}$ as controlled by the blending weights $\Omega$. More details about the SMPL model can be found in~\cite{loper2015smpl}.

Building upon the above GMM based probabilistic human model and under the i.i.d. assumption, we define the overall objective function as minimizing the negative log-likelihood for all the vertices of all the input scans from the training set, 
\begin{equation}
\small
\label{eq:energyEq}
E(\Theta,\sigma^2)  =  -\sum_{n=1}^{N} \log \left( (1-\mu)\sum_{m=1}^{M}    \pi_{mn} p(\bm{v}_n|m)  +  \frac{\mu}{N} \right), 
\end{equation}
Inspired by EM based optimization procedure to solve the above energy function iteratively~\cite{EM77}, we design a network with a probability association module in the forward pass to update the posterior distribution $p(m|v_n)$ and resolve the human model parameters in the backward training. 

%For global pose estimation, it is hard to train a neural network to directly predict the rotation matrix $R \in \mathbb{R}^{3 \times 3}$. 

\subsection{Network Structure}
\label{sec:net}
The overall network structure is shown in Fig.~\ref{fig:pipeline}. Basically, the network takes the point cloud as input and predicts the parameters of the template model. We use PointNet++~\cite{qi2017pointnetplusplus} to regress the parameters. In the forward pass, we propose a probabilistic correspondence association module to update the posterior probability of the the point cloud conditioned on the current predicted human model from the network; from that we define a novel loss function which minimizes the discrepancy between the deformed template and the point cloud conditioned on the posterior probability. In the backward pass, the network gets trained with the proposed loss function and the network prediction for parameters of the template human models gets updated afterwards. 
To illustrate the updating process, in Fig.~\ref{fig:pipeline} we mark three vertices over the input scan and show the updated soft correspondence association using the corresponding color over the human model. While the training proceeds, the correspondence map gets more localised and finally we get the deformed human model that fits closely to the input point cloud.

%The network is trained from scratch and in the forward pass the probabilistic correspondences association is updated by computing the posterior distribution of the input point cloud given the current predicted human model, and the network is trained with our proposed unsupervised loss to update the human model parameters via back-propagation.  

%Inspired by EM based optimization procedure, in the forward pass of network, we compute the correspondences association as the posterior probability distribution of the vertices on the template model given the input point cloud,
\subsection{Probabilistic Correspondence Association}
Similar to the E-step of the EM optimization procedure, given the parameters predicted by the network from the previous iteration, we update the posterior distribution of the vertices on the human model conditioned on the input point cloud by
\begin{equation}
\label{eq:E}
p^{(k)}(m|\bm{v}_n(\Theta)) = \frac{\exp \left(-\frac{||\bm{v}_n-M_m(\Theta^{(k-1)})||^2}{2\sigma_{k-1}^2} \right)} {\sum_{i=1}^{M} \exp \left(-\frac{||\bm{v}_n-M_i(\Theta^{(k-1)})||^2}{2\sigma_{k-1}^2}\right)+c},
\end{equation}
where $c = \big(2\pi(\sigma_{k-1})^2 \big)^{\frac{\bm{3}}{2}} \frac{1-\mu}{\mu} \frac{M}{N}$, and $k$ denotes the training iteration.
In our current scenario, the posterior distribution computation corresponds to the correspondence matching between the template model and the input point cloud. Compared with LoopReg~\cite{bhatnagar2020loopreg} which uses a diffusion field to differentiate the correspondence matching operation, our probabilistic correspondences association is naturally differential without including extra effort.
%To simplify the expression, we let $p^{old}$ stand for $p^{old}(\bm{t}_m(\phi)|\bm{s}_n)$ in the following sections.

\subsection{Self-supervised Loss}
Taking the computed posterior distribution into Eq.~\ref{eq:energyEq}, we update the human model parameters $\Theta$ by minimizing the following complete negative log-likelihood function.
\begin{equation} \label{eq:M}
\centering
\small
\begin{split}
Q(\Theta,\sigma^2) & =\frac{1}{2\sigma^2} \sum_{n,m=1}^{N,M}  p^{(k)}(m|v_n(\Theta)) \left\|v_n - M_m(\Theta) \right\|^2  \\ &   + \frac{3}{2} N_{\bm{P}} \text{log}\sigma^2
\end{split}
\end{equation}
% \vspace{-5pt}
% \begin{equation}
% \centering
% \small
% N_{\bm{P}} = \sum_{n=1}^{N}\sum_{m=1}^{M}  p^{(k)}(m|\bm{v}_n(\Theta)).
% \end{equation}
%The optimization has been validated to get converged after several iterations.
where $N_{\bm{P}} = \sum_{n=1}^{N}\sum_{m=1}^{M}  p^{(k)}(m|\bm{v}_n(\Theta))$. In the above function (Eq.~\ref{eq:M}), for the typical EM optimization process, $\Theta$ and $\sigma^2$ are supposed to be updated in the M-step. In our case, taking the point cloud as input, the network predicts the parameters $\Theta$ of the human model. But $\sigma$ is not controlled directly by the input point cloud, instead, it is treated as a hyper-parameter in the loss function. We develop a strategy to update $\sigma$ during the training iteration.
%In our scenario, different from the parameters in the template parametric human model, $\sigma$ is not controlled directly by the input point cloud. Instead we treat it as a hyper-parameter and develop a strategy to update it during the training procedure. 
Therefore, neglecting the second term in Eq.~\ref{eq:M} which has no dependency with respect to the model parameters, we define our self-supervised loss function as
\begin{equation} 
\label{eq:unsup}
\begin{aligned}
L_{unsup}(\Theta) =\frac{1}{2\sigma^2} \sum_{n,m=1}^{N,M}  p^{(k)}(m|\bm{v}_n(\Theta)) ||\bm{v}_n - M_m(\Theta)||^2.
\end{aligned}
\end{equation}
To prevent arbitrary poses and shapes, our training loss also include three regularization terms for the predicted human shape and pose parameters~\cite{bogo2016keepsmplify, kolotouros2019SPIN},
\begin{equation}
\label{eq:reg}
L_{reg}(\theta, \beta) = \lambda_{\theta} L_{\theta}(\theta) + \lambda_{a} L_{a}(\theta) + \lambda_{\beta} L_{\beta}(\beta), 
\end{equation}
where $L_{\theta}(\theta)$ is a mixture of Gaussian pose prior trained with shapes fitted on marker data~\cite{loper2014mosh}, $L_{a}(\theta)$ is a pose prior penalizing unnatural rotations of elbows and knees, while $L_{\beta}(\beta)$ is a quadratic penalty on the shape coefficients.

Finally, the overall loss function to train the network is expressed as
\begin{equation} 
L = L_{unsup}(\Theta) + L_{reg}(\theta, \beta). 
\end{equation}

\textbf{Update of $\sigma^2$.} In Eq.~\ref{eq:GMM}, the hyper-parameter $\sigma^2$ represents the variance of the Gaussian distribution controlling the certainty of the matching between the human model and input point cloud. Therefore, in the early iterations $\sigma$ is supposed to be larger since the initial human model has large distance to the input point cloud and we have great matching uncertainty. $\sigma$ gets smaller as the iteration proceeds. When $\sigma \to 0$, the posterior distribution (Eq.~\ref{eq:E}) will be close to a one-hot vector and our proposed loss function will approximate the classical Chamfer distance. In this way, we can see our proposed loss function as a generalized Chamfer distance with soft correspondences.

In the typical EM optimization process, $\sigma^2$ is updated via the following equation (Eq.~\ref{eq:sigma}), which is derived by setting the derivatives of $Q(\Theta,\sigma^2)$ to zero with respect to $\sigma^2$,
\begin{equation}
\label{eq:sigma}
\sigma^{2} = \frac{1}{3N_P} \sum_{n=1,m=1}^{N,M} p^{(k)}(m|v_n(\Theta)) ||\bm{v}_n - M_m(\Theta)||^2.
\end{equation}
As shown in the above equation, the update of $\sigma^2$ relies on the posterior probability of previous iteration which requires extra efforts to memorize and maintain it. In addition, different from traditional EM optimization, $\sigma^2$ will also get affected by the current status of the network. Therefore, we re-compute the posterior probability matrix with a $\sigma_{itr}$ decreasing linearly along with the training epoch and then update the current $\sigma^2$ via Eq.~\ref{eq:sigma}.

%\subsection{Global Pose Estimation} 
%Our network predicts the human model parameters which include shape, local pose and global pose parameters. For global pose estimation, it is hard to train a neural network to directly predict the rotation matrix $\mathbf{R} \in \mathbb{R}^{3 \times 3}$. Instead of using Euler angles to represent the global rotation~\cite{bhatnagar2020IPnet,bhatnagar2020loopreg}, in this paper we employ the 6D rotation representation which has been proved to be continuous in real Euclidean spaces and more suitable for learning~\cite{zhou2019continuity, xu2019disn}. Mathematically, we use the vector $\mathbf{b} = ( \mathbf{b_x} , \mathbf{b_y} ) \in \mathbb{R}^6 $, $\mathbf{b_x} \in \mathbb{R}^3 $, $\mathbf{b_y} \in \mathbb{R}^3$ to present the rotation, from which the rotation matrix $\mathbf{R} = (\mathbf{R_x}, \mathbf{R_y}, \mathbf{R_z})^T \in \mathbb{R}^{3 \times 3}$ can be obtained by 
%\begin{equation}
%    \mathbf{R_x} = N(\mathbf{b_x}), \quad
%    \mathbf{R_z} = N(\mathbf{R_x} \times \mathbf{b_y}), \quad 
%    \mathbf{R_y} = \mathbf{R_z} \times \mathbf{R_x}, 
%\end{equation}
%where $\mathbf{R_x}, \mathbf{R_y}, \mathbf{R_z} \in \mathbb{R}^3$, $N$ is a normalization function, "$\times$" means cross product.

%In the experimental section, we validate the effectiveness of using 6D representation in our framework. 

\subsection{Compatible with Complete and Incomplete Point Clouds} 
Previous methods on self-supervised human reconstruction or correspondence matching are usually designed to work with relatively complete point cloud~\cite{bhatnagar2020loopreg,wang2021locallyPTF,li2019lbsself}. To deal with in-complete point cloud, especially for the point cloud acquired from depth map where a great portion of the data is invisible, the existing works always rely on large human dataset with strong supervision~\cite{bhatnagar2020IPnet}. 
To be different, our proposed self-supervised method can naturally work with incomplete point cloud and the success comes from our implicit correspondence association in which we do not need to predict one-to-one correspondence between the template model and the input point cloud. %Basically, we also train the network from scratch and when it converges, the 
We demonstrate the effectiveness of our method on human shape and pose reconstruction from a depth map in the experiments.

\subsection{Implementation details}
Our network takes a point cloud of 2048 vertices as input which is sampled from the 3D human scans with farthest point sampling. In our loss function(Eq.~\ref{eq:reg}), $\lambda_{\theta}$, $\lambda_{a}$ and $\lambda_{\beta}$ is set as 20.0, 225.0 and 25.0 respectively. $\sigma_{itr}^2$ is initialized as 0.1.
We further improve the reconstruction with an instance-level optimization. Specifically, starting from the parametric model predicted from our network, we minimize the objective function defined in Eq.~\ref{eq:GMM} with EM optimization.

\section{Experiments}
\label{sec:exp}
\subsection{Datasets}
In this paper, we have considered three public datasets on human modeling: CAPE~\cite{ma2020cape}, FAUST~\cite{Bogo2014faust} and CMU Panoptic PointCloud Dataset~\cite{joo2017panoptic}. 
The \textbf{CAPE} dataset provides 148,584 pairs of scans under clothing and registered ground truth body shapes for 15 subjects of different genders.
The \textbf{FAUST} dataset consists of 100 training and 200 testing scans. They may include noise and have holes, typically missing part of the feet. In this paper, we have not used the training set to train our network, instead we only evaluated the correspondence matching results on the test set and made comparison with other methods.
The \textbf{CMU Panoptic PointCloud} Dataset is captured by 10 Synchronized Kinects. Compared with other 3D point cloud datasets, the captured human surface is quite noisy and contains large amount of vertices which do not belong to the human surface. For example, there are sequences of human subject playing musical instruments. Among the captured sequences, we choose the sequences containing single human subject that have ground-truth 3D joints for testing. 

\textbf{Training Set.}
We have split the CAPE dataset into training and testing set. Specifically, we use 12 subjects for training and 3 subjects for testing. We also subsampled the recordings by a factor of 5. The final training set consists of 26,004 frames while the validation set consists of 3,965 frames. We generated input point clouds by sampling 2,048 points on the surfaces of the clothed meshes and add Gaussian noise of zero mean and 1mm standard-deviation. 
%In this paper our network was trained on the CAPE training set but evaluated directly on other datasets without further fine-tuning. 
%we choose single human action sequence with 3D joints ground-truth,
%Kungfu Tracking dataset~\cite{guo2015robustkungfu}

%\subsection{Comparison Methods}
%SOTA supervised methods IPNet~\cite{bhatnagar2020IPnet} and PTF~\cite{wang2021locally}
\subsection{Comparisons on Human Shape and Pose Reconstruction}
\textbf{Comparison Methods.}
We compared with three state-of-the-art methods on human model reconstruction from point cloud, namely 3D-CODED~\cite{groueix20183dcoded}, IPNet~\cite{bhatnagar2020IPnet} and PTF~\cite{wang2021locallyPTF}, all of which are supervised approaches trained with groundtruth annotation. 
%For example, PTF~\cite{wang2021locallyPTF} is trained on CAPE dataset.
Both IPNet and PTF exploited implicit representations and SMPL based parametric model for surface fitting. 3D-CODED also adopted the SMPL template and directly regressed the deformation and predicted the deformed template model. For fair comparison, we compute the error of the predicted SMPL model. Besides, the 3D-CODED method cannot apply on the point cloud since they require a 3D mesh as input. Therefore, 3D-CODED method was not evaluated on CMU Panoptic PointCloud Dataset.

\begin{figure}[!ht]
    \centering
    \includegraphics[width=0.85\columnwidth]{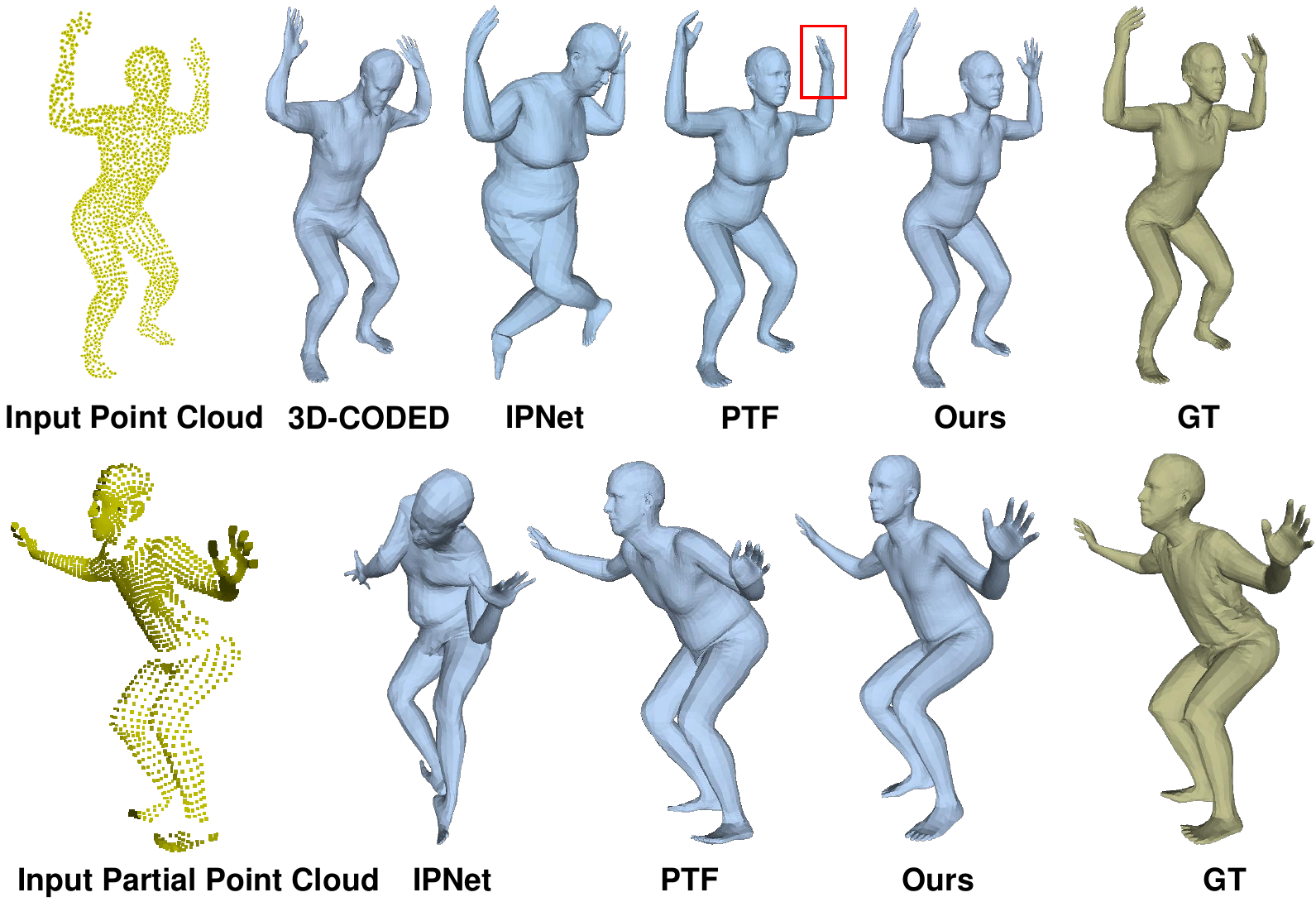}
    \caption{Qualitative evaluation on CAPE. From left to right, we show the input point cloud, SMPL mesh reconstructed by 3D-CODED~\cite{groueix20183dcoded}, IPNet~\cite{bhatnagar2020IPnet}, PTF~\cite{wang2021locallyPTF}, our result and the ground-truth clothed mesh~\cite{ma2020cape}.}
    \vspace{-5pt}
    \label{fig:pccmp}
\end{figure}

%In detail, IPNet(Implicit Part Network)~\cite{bhatnagar2020IPnet}combines detail-rich implicit functions and parametric representations to reconstruct 3D models of people. They jointly predict the outer 3D surface of the dressed person, the inner body surface, and the semantic correspondences to a parametric body model.
%Built upon implicit representations, PTF~\cite{wang2021locallyPTF} extends IPNet~\cite{bhatnagar2020IPnet} by regressing piecewise transformation fields (PTF) from local point cloud features, a set of functions that learn 3D translation vectors to map any query point in posed space to its correspond position in rest-pose space, from which the SMPL pose parameters are estimated. 
% \begin{figure}[!ht]
%     \centering
%     \includegraphics[width=0.9999\columnwidth]{figs/results_cmp_CAPE.pdf}
%     \caption{Qualitative evaluation on CAPE. From left to right, we show the input point cloud, SMPL mesh reconstructed by 3D-CODED~\cite{groueix20183dcoded}, IPNet~\cite{bhatnagar2020IPnet}, PTF~\cite{wang2021locallyPTF}, our result and the ground-truth clothed mesh~\cite{ma2020cape}.}
%     \label{fig:pccmp}
% \end{figure}
% \begin{figure}[!h]
%     \centering
%     \includegraphics[width=0.9999\columnwidth]{figs/results_cmp_CAPE_depth.pdf}
%     \caption{Qualitative evaluation on CAPE of incomplete data. From left to right, we show the input point cloud, SMPL mesh reconstructed by IPNet~\cite{bhatnagar2020IPnet}, ours and the ground-truth clothed mesh~\cite{ma2020cape}.}
%     \label{fig:depthcmp}
% \end{figure}

\begin{figure*}[!h]
    \centering
    \includegraphics[width=0.75\textwidth]{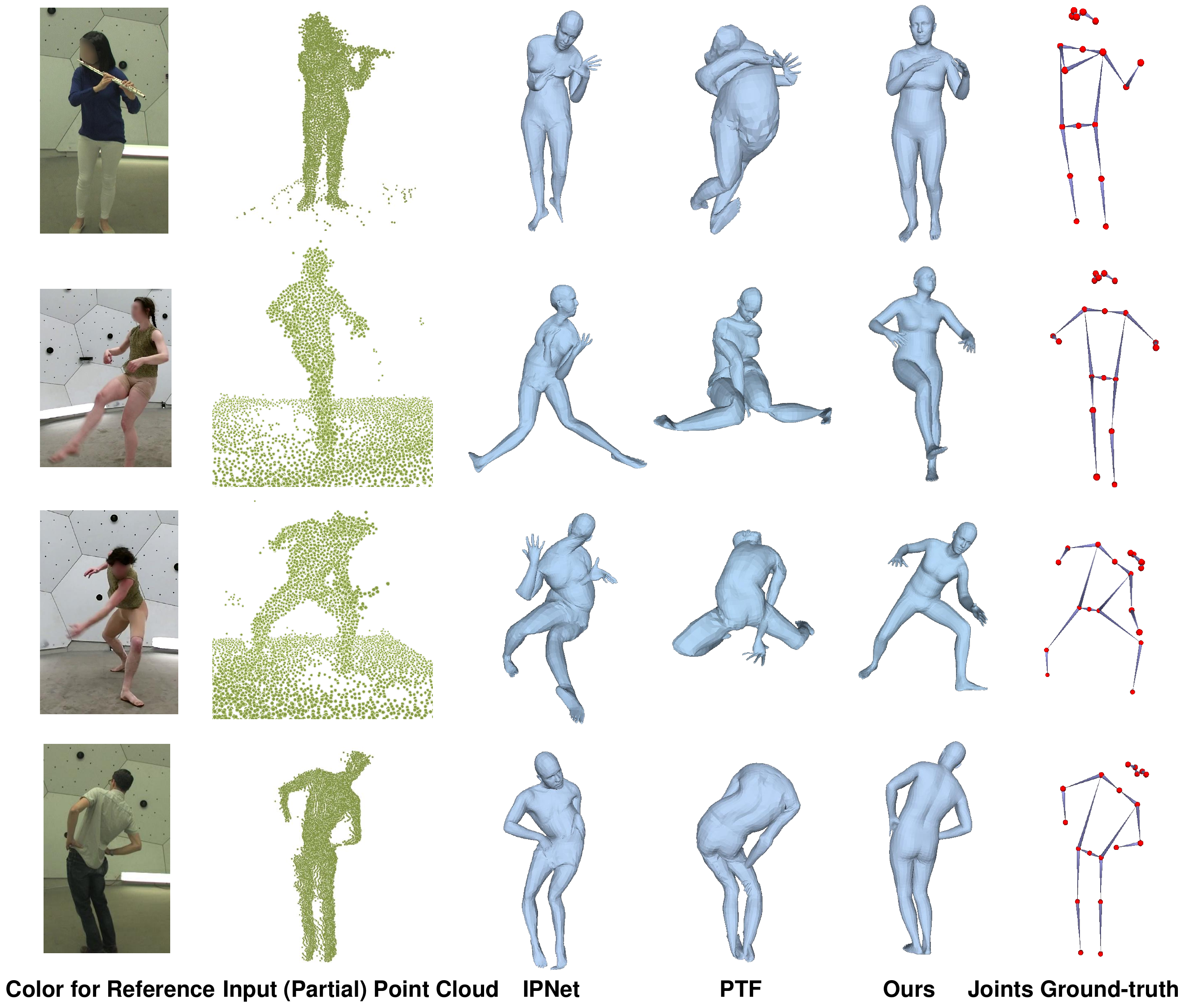}
    \caption{Qualitative evaluation on CMU Panoptic. From left to right, we show the reference color image, input point clouds, SMPL mesh reconstructed by IPNet~\cite{bhatnagar2020IPnet}, PTF~\cite{wang2021locallyPTF} and ours. The last column is the ground-truth 3D joints. \textbf{The color image is not used as input but only for visualization purpose.}}
    \label{fig:cmupc}
\end{figure*}

\subsubsection{Evaluation on CAPE Dataset}

\textbf{Evaluation Metrics.}
First, the vertex-to-vertex (V2V) error is computed as the distance between the vertex on the predicted SMPL model and the corresponding vertex on the groundtruth.
Besides, we also report Chamfer distance (CD) which is a common evaluation indicator for surface registration. It is computed as the mean distance of points on the reconstructed human mesh to their nearest neighbors on the ground truth human model.

To validate the effectiveness of our method for incomplete point cloud, we have generated depth maps for the models in CAPE dataset by rendering the human mesh into randomly generated camera coordinates rotating around the human subject. 

In Tab.~\ref{tab:dataset-compare-cape}, we demonstrate the evaluation results of 3D-CODED~\cite{groueix20183dcoded}, IPNet~\cite{wang2021locallyPTF}, PTF~\cite{bhatnagar2020IPnet} and ours. For those comparison methods, we use the pre-trained network released by the authors. As shown in Tab.~\ref{tab:dataset-compare-cape}, our proposed methods even outperforms the existing supervised approaches with smallest V2V and CD error for both complete point cloud and incomplete point cloud indicated as Depth. Since 3D-CODED cannot directly extend to incomplete point cloud, the corresponding error is not reported in the table.
In addition to quantitative evaluation, we also visualize some sampled results in Fig.~~\ref{fig:pccmp}. More visualization results can be found in the supplementary.
\begin{table}[!ht]
    \centering
    \begin{tabular}{c|cccc}
    \bottomrule 
    \hline
    \multirow{2}{*} {Method} & \multicolumn{2}{c}{Point Cloud} &  \multicolumn{2}{c}{Depth}  \\
    \cline{2-5}
    & V2V$\downarrow$ & CD$\downarrow$ & V2V$\downarrow$ & CD$\downarrow$
    \\
    \hline
        3D-CODED~\cite{groueix20183dcoded}  & 35.7 & 20.4  & - & - \\
        PTF~\cite{wang2021locallyPTF}   & 23.1 & 14.8 & 52.3 & 40.6  \\
        IPNet~\cite{bhatnagar2020IPnet}  & 28.2 & 15.1 & 60.2 & 47.2 \\
        Ours  & \textbf{21.8} & \textbf{13.2} & \textbf{48.6} & \textbf{33.2}   \\
    % \bottomrule
    \hline \toprule 
    \end{tabular}
    \caption{Quantitative evaluation on CAPE dataset. unit: mm}
    \label{tab:dataset-compare-cape}\vspace{-5pt}
\end{table}

\subsubsection{Evaluation on CMU Panoptic Dataset}

\textbf{Evaluation Metrics.}
For the CMU Panoptic dataset, we do not have the groundtruth SMPL for the captured point cloud. Instead, the groundtruth 3D joints are provided. Therefore, we use Mean Per Joint Position Error (MPJPE) as evaluation protocol which is computed as the average Euclidean distance of predicted joints to the ground-truth.
In addition, percent of correct keypoints (PCK) error is also reported where computed joint is considered correct if its distance to the groundtruth is within a certain threshold (100 mm in this paper).
\begin{table}
\centering
    \begin{tabular}{c|cccc}
    \bottomrule 
    \hline
    \multirow{2}{*} {Method} & \multicolumn{2}{c}{---Point Cloud---} &  \multicolumn{2}{c}{--------Depth--------}  \\
    \cline{2-5}
    & MPJPE$\downarrow$ & PCK$\uparrow$ & MPJPE $\downarrow$ & PCK$\uparrow$
    \\
    \hline
    IPNet~\cite{bhatnagar2020IPnet} & 43.8 & 73.2 & 56.8 & 57.2   \\
    \hline
    PTF~\cite{wang2021locallyPTF} & 41.1 & 75.2 & 52.4 & 63.7   \\
    \hline
    Ours  & \textbf{25.7} & \textbf{89.6} & \textbf{37.6} & \textbf{79.2}   \\
    % \bottomrule
    \hline \toprule 
    \end{tabular}
    \caption{Joints error on CMU Panoptic. Unit for MPJPE:  mm.}
    \vspace{-10pt}
    \label{tab:dataset-compare-cmu}
\end{table}

In Tab.~\ref{tab:dataset-compare-cmu}, we show the error of joints extracted from the predicted SMPL model and also report the error on both complete and incomplete point cloud. For the complete point cloud we used the point cloud merged from all the 10 Kinects and for incomplete point cloud we used the depth map randomly selected from those Kinects. As shown in Tab.~\ref{tab:dataset-compare-cmu}, the proposed method has outperformed the current SOTA methods in a big margin with smallest error with respect to MPJPE and PCK. Some sampled results are displayed in Fig.~\ref{fig:cmupc}, from which we can that the proposed method is robust to noisy input point cloud containing great outliers. On the contrary, the current SOTA methods failed to reconstruct plausible human models due to the great noise and outliers in the input point clouds. More visualization results can be found in the supplementary.

\subsection{Comparison on Correspondence Prediction}
\textbf{Comparison Methods.} 
We have compared with several existing approaches on correspondence matching. Specifically, FMNet~\cite{litany2017deepfunctional} and LoopReg~\cite{bhatnagar2020loopreg} can directly regress the correspondences while 3D-CODED~\cite{groueix20183dcoded}, LBS-AE~\cite{li2019lbsself} as well as our proposed approach use the predicted template model as the bridge to compute the correspondences between different scans. Among those methods, FMNet and 3D-CODED are supervised approaches while LoopReg and LBS-AE are unsupervised ones, but they still rely on supervised data to warm-up the training.

\textbf{Evaluation on FAUST Dataset.}
We have compared with previous approaches of correspondence matching on FAUST dataset~\cite{Bogo2014faust} with the results shown in Tab.~\ref{tab:faust_dataset}. 
We follow the evaluation metric in paper~\cite{bhatnagar2020loopreg} and report the correspondence error as Euclidean distance of the predicted correspondence to the groundtruth for both Inter-class and Intra-class. As shown in Tab.~\ref{tab:faust_dataset}, our proposed method has achieved the best performance on both cases and the matching error was reduced in a large margin especially for the Intra-class scenario.
\begin{table}[h]
    \small
    \centering
    \begin{tabular}{c|ccccc}
    \bottomrule \hline
        \footnotesize{Metrics} & \footnotesize{FMNet} & \footnotesize{LBS-AE} & \footnotesize{3D-CODED} & \small{LoopReg} & \footnotesize{Ours}\\
    % \midrule
    \hline
        \footnotesize{Inter-class$\downarrow$} & 48.3 & 40.8 & 28.7 & 26.6 & \textbf{22.38}  \\
        \footnotesize{Intra-class$\downarrow$} & 24.4 & 21.6 & 19.8 & 13.4 & \textbf{9.97} \\
    % \bottomrule
    \hline \toprule 
    \end{tabular}
    \caption{Correspondence evaluation on FAUST dataset. Our method clearly outperforms the state-of-the-art supervised~\cite{litany2017deepfunctional, groueix20183dcoded} and unsupervised~\cite{li2019lbsself, bhatnagar2020loopreg} approaches. FMNet~\cite{litany2017deepfunctional} LBS-AE~\cite{li2019lbsself} 3D-CODED~\cite{groueix20183dcoded}  LoopReg~\cite{bhatnagar2020loopreg}. The unit of the report error is mm. }
    \vspace{-5pt}
    \label{tab:faust_dataset}
\end{table}

\subsection{Ablation Study}
%We conduct ablation studies on several key components of the proposed method with the CAPE dataset.
In this section, we conducted ablation studies on several key components of the proposed method. We implemented the following ablation studies on the CAPE dataset.
\begin{table}[b]
    \centering
    \begin{tabular}{c|ccccc}
    \bottomrule 
    \hline
    \multirow{4}{*} {Noise} & \multicolumn{5}{c}{Standard Deviation} \\
    \cline{2-6}
     & 0 & 5 & 10 & 20 & 50 \\
     \cline{2-6}
    & \multicolumn{5}{c}{Outlier Percentage} \\
    \cline{2-6}
    & $0\%$  & $5\%$ & $10\%$  & $15\%$ & $20\%$ \\
    \hline
    PTF~\cite{wang2021locallyPTF}   V2V$\downarrow$ & 23.1  & 52.3  & 59.6  & 64.5 & 71.4  \\
    \hline
    PTF~\cite{wang2021locallyPTF}  CD$\downarrow$ & 14.8
    & 26.7  & 29.0 & 34.4  & 38.8  \\
    \hline
   ours V2V$\downarrow$ & 21.8 & 22.9 & 26.7 & 28.2 & 32.5 \\
    \hline
    ours CD$\downarrow$ & 13.2 & 14.5 & 15.8 & 16.7 & 19.2 \\
    % \bottomrule
    \hline \toprule 
    \end{tabular}
    \caption{Ablation study on surface noise. unit(mm)}
    \vspace{-10pt}
    \label{tab:noise}
\end{table}

\textbf{Robustness to Noise.}
We have added both Gaussian noise and random outliers for the models in the CAPE test set. In Tab.~\ref{tab:noise}, we can see that while increasing the standard deviation of the Gaussian noise and the percentage of random outliers, compared with PTF~\cite{wang2021locallyPTF} that has not considered the noise, the V2V and Chamfer errors of the predicted human surface from our proposed method had a rather small increase. For example, after adding Gaussian noise with 5mm standard deviation and $5\%$ percentage of outliers, the V2V and Chamfer errors of our predicted human model only increased by 1.1mm and 1.3mm respectively. On the contrary, the errors of PTF method have increased greatly.

\begin{figure}[!h]
\centering
\includegraphics[width=0.9\columnwidth]{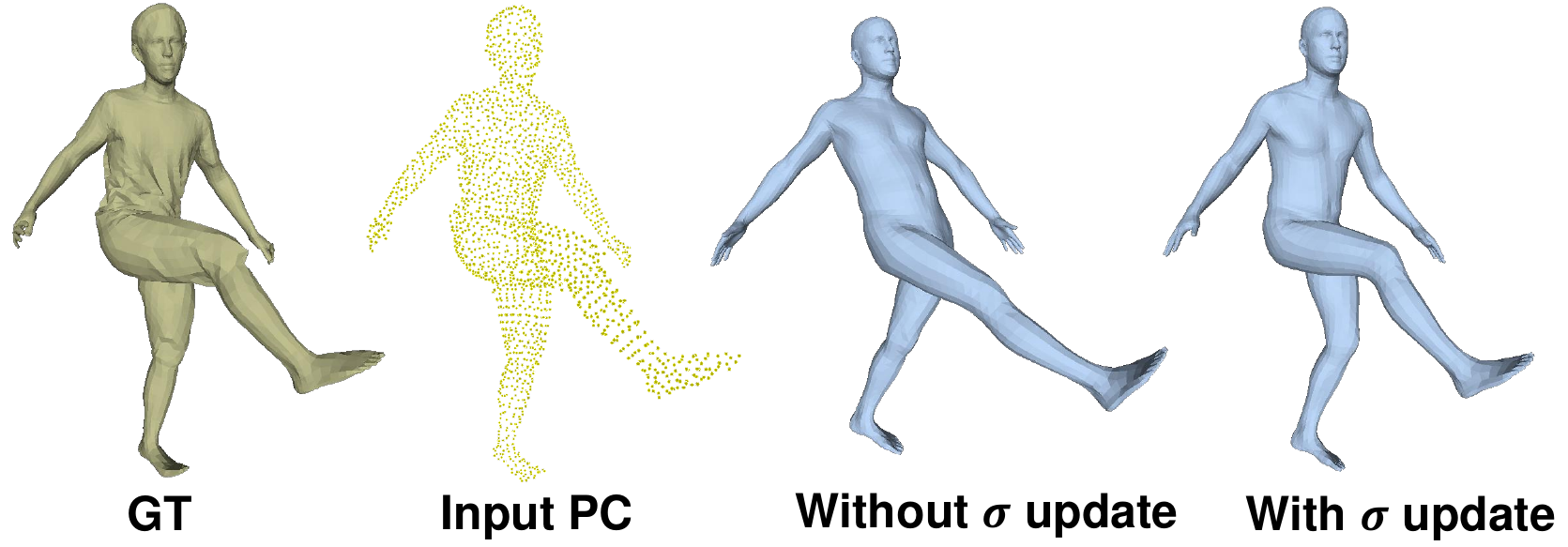}
\caption{Ablation study on $\sigma^2$ update. From left to right, we show the ground-truth clothed meshes~\cite{ma2020cape} from CAPE dataset, input point clouds, reconstructed human model when trained using fixed $\sigma^2$, and trained using progressively updated $\sigma^2$.}
\label{fig:sigmacmp}
\end{figure}

\textbf{Update of $\sigma^2$.}
In Fig.~\ref{fig:sigmacmp}, we show sampled comparison results of the reconstructed human model w/o updating $\sigma^2$ during training. Trained with small $\sigma^2$, the network will have similar performance as using Chamfer distance where it is difficult for the training to converge. On the other hand, as demonstrated in Fig.~\ref{fig:sigmacmp}, with large $\sigma^2$ the network lacks the ability to precisely fit to the input scan. On the contrary, with our proposed updating strategy, we can have better fitting to the input point cloud. In Tab.~\ref{tab:sigmaAB} we show the quantitative results of the reconstructed human model as compared with using fixed $\sigma^2$ and the progressively updated $\sigma^2$ with our proposed update strategy.
\begin{table}[!]
    \centering
    \begin{tabular}{c|cccc}
    \bottomrule \hline
        $\sigma^2$ & 0.05 & 0.1 & 0.15 & with updated $\sigma^2$\\
    % \midrule
    \hline
       V2V$\downarrow$   & 26.7 & 25.9 & 27.4 & \textbf{21.8} \\
       CD$\downarrow$  & 14.6 & 14.5 & 14.8 & \textbf{13.2} \\
    % \bottomrule
    \hline \toprule 
    \end{tabular}
    \caption{Ablation study of updating $\sigma^2$. unit(mm).}
    \label{tab:sigmaAB}
\end{table}

\textbf{The importance of self-supervised loss.} Although we could pre-train the network on synthetic datasets with fully supervised loss on the human parameters, it is always good to have self-supervised loss with which we could either fine-tune on real data or even train the network from scratch without pre-training on synthetic dataset. In this section, we conduct an ablation study for the proposed self-supervised loss to validate its importance and effectiveness. Basically, the network was pre-trained on a synthetic dataset (CAPE dataset~\cite{ma2020cape}) and we evaluate the performance of the network with and without fine-tuning on the real dataset (CMU Panoptic dataset~\cite{joo2017panoptic}) using our proposed self-supervised loss. The quantitative evaluations are shown in Tab.~\ref{tab:selfLossAb}, from which we can see that the human reconstruction performance gets improved greatly with the joints error reduced in a large margin after fine-tuning using the self-supervised loss.
\begin{table}[!ht]
    \centering
    \begin{tabular}{c|ccccc}
    \bottomrule \hline
        Method &  MPJPE$\downarrow$ & PCK$\uparrow$\\
    % \midrule
    \hline
        without fine-tuning & 36.6 & 68.3  \\
        fine-tuned with self-supervised loss & \textbf{25.7} & \textbf{89.6}  \\
    % \bottomrule
    \hline \toprule 
    \end{tabular}
    \caption{Ablation study of the proposed self-supervised loss. The unit of MPJPE is mm.}
    \label{tab:selfLossAb}
\end{table}

\section{Conclusion and Limitation}
In this paper, a novel self-supervised approach was proposed to reconstruct human shape and pose from noisy input point clouds. Instead of explicitly predicting or regressing the correspondences for each vertice of the point cloud, which has no tolerance to outliers, we have adopted GMM to model the input point cloud, and implicitly encoded the correspondences with probabilistic association. A novel loss function was proposed to train the network in a fully self-supervised manner, which is also robust to a significant amount of outliers. We have conducted evaluation on several public datasets and outperformed both supervised and unsupervised state-of-the-art methods in a big margin especially for noisy point clouds from real captured datasets. 

\textbf{Limitations.} As a limitation, we have not explicitly addressed the collision problem in this paper which makes it difficult to precisely reconstruct the human surface when the human body has close self-interaction. As a future work, we could include body part information to resolve this issue.

%%%%%%%%% REFERENCES
{\small
\bibliographystyle{ieee_fullname}
\bibliography{mybib}

\begin{thebibliography}{10}\itemsep=-1pt

\bibitem{aubry2011wave}
Mathieu Aubry, Ulrich Schlickewei, and Daniel Cremers.
\newblock The wave kernel signature: A quantum mechanical approach to shape
  analysis.
\newblock In {\em 2011 IEEE international conference on computer vision
  workshops (ICCV workshops)}, pages 1626--1633. IEEE, 2011.

\bibitem{besl1992icp}
Paul~J Besl and Neil~D McKay.
\newblock Method for registration of 3-d shapes.
\newblock {\em IEEE Transactions on Pattern Analysis and Machine Intelligence},
  14(2):239--256, 1992.

\bibitem{bhatnagar2020IPnet}
Bharat~Lal Bhatnagar, Cristian Sminchisescu, Christian Theobalt, and Gerard
  Pons-Moll.
\newblock Combining implicit function learning and parametric models for 3d
  human reconstruction.
\newblock In {\em Proceedings of the European Conference on Computer Vision
  (ECCV)}, pages 311--329, 2020.

\bibitem{bhatnagar2020loopreg}
Bharat~Lal Bhatnagar, Cristian Sminchisescu, Christian Theobalt, and Gerard
  Pons-Moll.
\newblock Loopreg: Self-supervised learning of implicit surface
  correspondences, pose and shape for 3d human mesh registration.
\newblock In {\em Advances in neural information processing systems}, pages
  12909--12922, 2020.

\bibitem{biggs2020left}
Benjamin Biggs, Oliver Boyne, James Charles, Andrew Fitzgibbon, and Roberto
  Cipolla.
\newblock Who left the dogs out? 3d animal reconstruction with expectation
  maximization in the loop.
\newblock In {\em European Conference on Computer Vision}, pages 195--211.
  Springer, 2020.

\bibitem{bogo2016keepsmplify}
Federica Bogo, Angjoo Kanazawa, Christoph Lassner, Peter Gehler, Javier Romero,
  and Michael~J Black.
\newblock Keep it smpl: Automatic estimation of 3d human pose and shape from a
  single image.
\newblock In {\em European conference on Computer Vision (ECCV)}, pages
  561--578. Springer, 2016.

\bibitem{Bogo2014faust}
Federica Bogo, Javier Romero, Matthew Loper, and Michael~J. Black.
\newblock {FAUST}: Dataset and evaluation for {3D} mesh registration.
\newblock In {\em Proceedings of the IEEE Conference on Computer Vision and
  Pattern Recognition}, pages 3794--3801, 2014.

\bibitem{bogo2017dynamicfaust}
Federica Bogo, Javier Romero, Gerard Pons-Moll, and Michael~J Black.
\newblock Dynamic faust: Registering human bodies in motion.
\newblock In {\em Proceedings of the IEEE conference on computer vision and
  pattern recognition}, pages 6233--6242, 2017.

\bibitem{bronstein2010scale}
Michael~M Bronstein and Iasonas Kokkinos.
\newblock Scale-invariant heat kernel signatures for non-rigid shape
  recognition.
\newblock In {\em 2010 IEEE Computer Society Conference on Computer Vision and
  Pattern Recognition}, pages 1704--1711. IEEE, 2010.

\bibitem{browne2011model}
Ryan~P Browne, Paul~D McNicholas, and Matthew~D Sparling.
\newblock Model-based learning using a mixture of mixtures of gaussian and
  uniform distributions.
\newblock {\em IEEE Transactions on Pattern Analysis and Machine Intelligence},
  34(4):814--817, 2011.

\bibitem{chibane2020implicit}
Julian Chibane, Thiemo Alldieck, and Gerard Pons-Moll.
\newblock Implicit functions in feature space for 3d shape reconstruction and
  completion.
\newblock In {\em Proceedings of the IEEE/CVF Conference on Computer Vision and
  Pattern Recognition}, pages 6970--6981, 2020.

\bibitem{EM77}
Arthur~P Dempster, Nan~M Laird, and Donald~B Rubin.
\newblock Maximum likelihood from incomplete data via the em algorithm.
\newblock {\em Journal of the Royal Statistical Society: Series B
  (Methodological)}, 39(1):1--22, 1977.

\bibitem{dou2016fusion4d}
Mingsong Dou, Sameh Khamis, Yury Degtyarev, Philip Davidson, Sean~Ryan Fanello,
  Adarsh Kowdle, Sergio~Orts Escolano, Christoph Rhemann, David Kim, Jonathan
  Taylor, et~al.
\newblock Fusion4d: Real-time performance capture of challenging scenes.
\newblock {\em ACM Transactions on Graphics (ToG)}, 35(4):1--13, 2016.

\bibitem{eisenberger2020deep}
Marvin Eisenberger, Aysim Toker, Laura Leal-Taix{\'e}, and Daniel Cremers.
\newblock Deep shells: Unsupervised shape correspondence with optimal
  transport.
\newblock In {\em Advances in neural information processing systems}, pages
  10491--10502, 2020.

\bibitem{feng2021recurrent}
Wanquan Feng, Juyong Zhang, Hongrui Cai, Haofei Xu, Junhui Hou, and Hujun Bao.
\newblock Recurrent multi-view alignment network for unsupervised surface
  registration.
\newblock In {\em Proceedings of the IEEE/CVF Conference on Computer Vision and
  Pattern Recognition}, pages 10297--10307, 2021.

\bibitem{ginzburg2020cyclic}
Dvir Ginzburg and Dan Raviv.
\newblock Cyclic functional mapping: Self-supervised correspondence between
  non-isometric deformable shapes.
\newblock In {\em European Conference on Computer Vision}, pages 36--52.
  Springer, 2020.

\bibitem{groueix20183dcoded}
Thibault Groueix, Matthew Fisher, Vladimir~G Kim, Bryan~C Russell, and Mathieu
  Aubry.
\newblock 3d-coded: 3d correspondences by deep deformation.
\newblock In {\em Proceedings of the European Conference on Computer Vision
  (ECCV)}, pages 230--246, 2018.

\bibitem{horaud2010rigid}
Radu Horaud, Florence Forbes, Manuel Yguel, Guillaume Dewaele, and Jian Zhang.
\newblock Rigid and articulated point registration with expectation conditional
  maximization.
\newblock {\em IEEE Transactions on Pattern Analysis and Machine Intelligence},
  33(3):587--602, 2010.

\bibitem{jiang2019skeleton}
Haiyong Jiang, Jianfei Cai, and Jianmin Zheng.
\newblock Skeleton-aware 3d human shape reconstruction from point clouds.
\newblock In {\em Proceedings of the IEEE/CVF International Conference on
  Computer Vision}, pages 5431--5441, 2019.

\bibitem{joo2017panoptic}
Hanbyul Joo, Tomas Simon, Xulong Li, Hao Liu, Lei Tan, Lin Gui, Sean Banerjee,
  Timothy Godisart, Bart Nabbe, Iain Matthews, et~al.
\newblock Panoptic studio: A massively multiview system for social interaction
  capture.
\newblock {\em IEEE Transactions on Pattern Analysis and Machine Intelligence},
  41(1):190--204, 2017.

\bibitem{kim2011blended}
Vladimir~G Kim, Yaron Lipman, and Thomas Funkhouser.
\newblock Blended intrinsic maps.
\newblock {\em ACM transactions on graphics (TOG)}, 30(4):1--12, 2011.

\bibitem{kolotouros2019SPIN}
Nikos Kolotouros, Georgios Pavlakos, Michael~J Black, and Kostas Daniilidis.
\newblock Learning to reconstruct 3d human pose and shape via model-fitting in
  the loop.
\newblock In {\em Proceedings of the IEEE/CVF International Conference on
  Computer Vision}, pages 2252--2261, 2019.

\bibitem{li2019lbsself}
Chun-Liang Li, Tomas Simon, Jason Saragih, Barnab{\'a}s P{\'o}czos, and Yaser
  Sheikh.
\newblock Lbs autoencoder: Self-supervised fitting of articulated meshes to
  point clouds.
\newblock In {\em Proceedings of the IEEE/CVF Conference on Computer Vision and
  Pattern Recognition}, pages 11967--11976, 2019.

\bibitem{li2008global}
Hao Li, Robert~W Sumner, and Mark Pauly.
\newblock Global correspondence optimization for non-rigid registration of
  depth scans.
\newblock {\em Computer graphics forum}, 27(5):1421--1430, 2008.

\bibitem{li2021posefusion}
Zhe Li, Tao Yu, Zerong Zheng, Kaiwen Guo, and Yebin Liu.
\newblock Posefusion: Pose-guided selective fusion for single-view human
  volumetric capture.
\newblock In {\em Proceedings of the IEEE/CVF Conference on Computer Vision and
  Pattern Recognition}, pages 14162--14172, 2021.

\bibitem{litany2017deepfunctional}
Or Litany, Tal Remez, Emanuele Rodola, Alex Bronstein, and Michael Bronstein.
\newblock Deep functional maps: Structured prediction for dense shape
  correspondence.
\newblock In {\em Proceedings of the IEEE International Conference on Computer
  Vision}, pages 5659--5667, 2017.

\bibitem{loper2014mosh}
Matthew Loper, Naureen Mahmood, and Michael~J Black.
\newblock Mosh: Motion and shape capture from sparse markers.
\newblock {\em ACM Transactions on Graphics (ToG)}, 33(6):1--13, 2014.

\bibitem{loper2015smpl}
Matthew Loper, Naureen Mahmood, Javier Romero, Gerard Pons-Moll, and Michael~J
  Black.
\newblock Smpl: A skinned multi-person linear model.
\newblock {\em ACM Transactions on Graphics (ToG)}, 34(6):1--16, 2015.

\bibitem{ma2021scale}
Qianli Ma, Shunsuke Saito, Jinlong Yang, Siyu Tang, and Michael~J Black.
\newblock Scale: Modeling clothed humans with a surface codec of articulated
  local elements.
\newblock In {\em Proceedings of the IEEE/CVF Conference on Computer Vision and
  Pattern Recognition}, pages 16082--16093, 2021.

\bibitem{ma2020cape}
Qianli Ma, Jinlong Yang, Anurag Ranjan, Sergi Pujades, Gerard Pons-Moll, Siyu
  Tang, and Michael~J Black.
\newblock Learning to dress 3d people in generative clothing.
\newblock In {\em Proceedings of the IEEE/CVF Conference on Computer Vision and
  Pattern Recognition}, pages 6469--6478, 2020.

\bibitem{ma2021power}
Qianli Ma, Jinlong Yang, Siyu Tang, and Michael~J Black.
\newblock The power of points for modeling humans in clothing.
\newblock In {\em Proceedings of the IEEE/CVF International Conference on
  Computer Vision}, pages 10974--10984, 2021.

\bibitem{marin2020farm}
Riccardo Marin, Simone Melzi, Emanuele Rodola, and Umberto Castellani.
\newblock Farm: Functional automatic registration method for 3d human bodies.
\newblock {\em Computer Graphics Forum}, 39(1):160--173, 2020.

\bibitem{mihajlovic2021leap}
Marko Mihajlovic, Yan Zhang, Michael~J Black, and Siyu Tang.
\newblock Leap: Learning articulated occupancy of people.
\newblock In {\em Proceedings of the IEEE/CVF Conference on Computer Vision and
  Pattern Recognition}, pages 10461--10471, 2021.

\bibitem{nguyen2011optimization}
Andy Nguyen, Mirela Ben-Chen, Katarzyna Welnicka, Yinyu Ye, and Leonidas
  Guibas.
\newblock An optimization approach to improving collections of shape maps.
\newblock {\em Computer Graphics Forum}, 30(5):1481--1491, 2011.

\bibitem{ovsjanikov2012functional}
Maks Ovsjanikov, Mirela Ben-Chen, Justin Solomon, Adrian Butscher, and Leonidas
  Guibas.
\newblock Functional maps: a flexible representation of maps between shapes.
\newblock {\em ACM Transactions on Graphics (TOG)}, 31(4):1--11, 2012.

\bibitem{pons2017clothcap}
Gerard Pons-Moll, Sergi Pujades, Sonny Hu, and Michael~J Black.
\newblock Clothcap: Seamless 4d clothing capture and retargeting.
\newblock {\em ACM Transactions on Graphics (ToG)}, 36(4):1--15, 2017.

\bibitem{qi2017pointnetplusplus}
Charles~R Qi, Li Yi, Hao Su, and Leonidas~J Guibas.
\newblock Pointnet++: Deep hierarchical feature learning on point sets in a
  metric space.
\newblock In {\em Advances in neural information processing systems}, pages
  5099--5108, 2017.

\bibitem{romero2017embodiedMANO}
Javier Romero, Dimitrios Tzionas, and Michael~J Black.
\newblock Embodied hands: Modeling and capturing hands and bodies together.
\newblock {\em ACM Transactions on Graphics (ToG)}, 36(6):1--17, 2017.

\bibitem{roufosse2019unsupervised}
Jean-Michel Roufosse, Abhishek Sharma, and Maks Ovsjanikov.
\newblock Unsupervised deep learning for structured shape matching.
\newblock In {\em Proceedings of the IEEE/CVF International Conference on
  Computer Vision}, pages 1617--1627, 2019.

\bibitem{saito2021scanimate}
Shunsuke Saito, Jinlong Yang, Qianli Ma, and Michael~J Black.
\newblock Scanimate: Weakly supervised learning of skinned clothed avatar
  networks.
\newblock In {\em Proceedings of the IEEE/CVF Conference on Computer Vision and
  Pattern Recognition}, pages 2886--2897, 2021.

\bibitem{solomon2016entropic}
Justin Solomon, Gabriel Peyr{\'e}, Vladimir~G Kim, and Suvrit Sra.
\newblock Entropic metric alignment for correspondence problems.
\newblock {\em ACM Transactions on Graphics (ToG)}, 35(4):1--13, 2016.

\bibitem{tombari2010unique}
Federico Tombari, Samuele Salti, and Luigi Di~Stefano.
\newblock Unique signatures of histograms for local surface description.
\newblock In {\em European conference on computer vision}, pages 356--369.
  Springer, 2010.

\bibitem{vestner2017product}
Matthias Vestner, Roee Litman, Emanuele Rodola, Alex Bronstein, and Daniel
  Cremers.
\newblock Product manifold filter: Non-rigid shape correspondence via kernel
  density estimation in the product space.
\newblock In {\em Proceedings of the IEEE conference on computer vision and
  pattern recognition}, pages 3327--3336, 2017.

\bibitem{wan2019self}
Chengde Wan, Thomas Probst, Luc~Van Gool, and Angela Yao.
\newblock Self-supervised 3d hand pose estimation through training by fitting.
\newblock In {\em Proceedings of the IEEE/CVF Conference on Computer Vision and
  Pattern Recognition}, pages 10853--10862, 2019.

\bibitem{wang2020sequential}
Kangkan Wang, Jin Xie, Guofeng Zhang, Lei Liu, and Jian Yang.
\newblock Sequential 3d human pose and shape estimation from point clouds.
\newblock In {\em Proceedings of the IEEE/CVF Conference on Computer Vision and
  Pattern Recognition}, pages 7275--7284, 2020.

\bibitem{wang2021learning}
Kangkan Wang, Guofeng Zhang, Huayu Zheng, and Jian Yang.
\newblock Learning dense correspondences for non-rigid point clouds with
  two-stage regression.
\newblock {\em IEEE Transactions on Image Processing}, 2021.

\bibitem{wang2021locallyPTF}
Shaofei Wang, Andreas Geiger, and Siyu Tang.
\newblock Locally aware piecewise transformation fields for 3d human mesh
  registration.
\newblock In {\em Proceedings of the IEEE/CVF Conference on Computer Vision and
  Pattern Recognition}, pages 7639--7648, 2021.

\bibitem{wei2016dense}
Lingyu Wei, Qixing Huang, Duygu Ceylan, Etienne Vouga, and Hao Li.
\newblock Dense human body correspondences using convolutional networks.
\newblock In {\em Proceedings of the IEEE Conference on Computer Vision and
  Pattern Recognition}, pages 1544--1553, 2016.

\bibitem{xu2019disn}
Qiangeng Xu, Weiyue Wang, Duygu Ceylan, Radomir Mech, and Ulrich Neumann.
\newblock Disn: Deep implicit surface network for high-quality single-view 3d
  reconstruction.
\newblock In {\em Advances in neural information processing systems}, page
  492–502, 2019.

\bibitem{ye2014real}
Mao Ye and Ruigang Yang.
\newblock Real-time simultaneous pose and shape estimation for articulated
  objects using a single depth camera.
\newblock In {\em Proceedings of the IEEE Conference on Computer Vision and
  Pattern Recognition}, pages 2345--2352, 2014.

\bibitem{yu2018doublefusion}
Tao Yu, Zerong Zheng, Kaiwen Guo, Jianhui Zhao, Qionghai Dai, Hao Li, Gerard
  Pons-Moll, and Yebin Liu.
\newblock Doublefusion: Real-time capture of human performances with inner body
  shapes from a single depth sensor.
\newblock In {\em Proceedings of the IEEE Conference on Computer Vision and
  Pattern Recognition}, pages 7287--7296, 2018.

\bibitem{zhou2019continuity}
Yi Zhou, Connelly Barnes, Jingwan Lu, Jimei Yang, and Hao Li.
\newblock On the continuity of rotation representations in neural networks.
\newblock In {\em Proceedings of the IEEE/CVF Conference on Computer Vision and
  Pattern Recognition}, pages 5745--5753, 2019.

\end{thebibliography}
}

\end{document}